\documentclass[letterpaper, 10 pt, conference]{ieeeconf}  

\IEEEoverridecommandlockouts                              

\usepackage{multirow, makecell}
\usepackage{hhline}
\usepackage{array, booktabs}
\usepackage{bm, amsmath}

\usepackage{graphicx} 
\usepackage{epsfig} 
\usepackage{float}
\usepackage[hidelinks]{hyperref}

\title{\LARGE \bf
Probabilistic Prediction of Vehicle Semantic Intention and Motion
}

\author{Yeping Hu, Wei Zhan and Masayoshi Tomizuka
\thanks{Y. Hu, W. Zhan and M. Tomizuka are with the Department of Mechanical Engineering, 
		University of California, Berkeley, CA 94720 USA 
		{\tt {[yeping\_hu, wzhan, tomizuka@berkeley.edu]}}}
}

\begin{document}

\maketitle
\thispagestyle{empty}
\pagestyle{empty}

\begin{abstract}
Accurately predicting the possible behaviors of traffic participants is an essential capability for future autonomous vehicles. The majority of current researches fix the number of driving intentions by considering only a specific scenario. However, distinct driving environments usually contain various possible driving maneuvers. Therefore, a intention prediction method that can adapt to different traffic scenarios is needed.
To further improve the overall vehicle prediction performance, motion information is usually incorporated with classified intentions.
As suggested in some literature, the methods that directly predict possible goal locations can achieve better performance for long-term motion prediction than other approaches due to their automatic incorporation of environment constraints. Moreover, by obtaining the temporal information of the predicted destinations, the optimal trajectories for predicted vehicles as well as the desirable path for ego autonomous vehicle could be easily generated. In this paper, we propose a Semantic-based Intention and Motion Prediction (SIMP) method, which can be adapted to any driving scenarios by using semantic-defined vehicle behaviors. It utilizes a probabilistic framework based on deep neural network to estimate the intentions, final locations, and the corresponding time information for surrounding vehicles. An exemplar real-world scenario was used to implement and examine the proposed method.

\end{abstract}

\section{Introduction}

Safety is the most fundamental aspect to consider for both human drivers and autonomous vehicles. Human drivers are capable of using past experience and intuitions to avoid potential accidents by predicting the behaviors of other drivers. However, some drivers have poor driving habits such as changing lanes without using turn signals, which adds difficulties for prediction. Moreover, human drivers might easily overlook dangerous situations due to limited concentration. Therefore, the Advanced Driver Assistance Systems (ADAS) should have the ability to simultaneously and accurately anticipate future behaviors of multiple traffic participants under various driving scenarios, which may then assure a safe, comfortable and cooperative driving experience. 

\begin{figure}[htbp]
	\centering
	\includegraphics[scale=0.35]{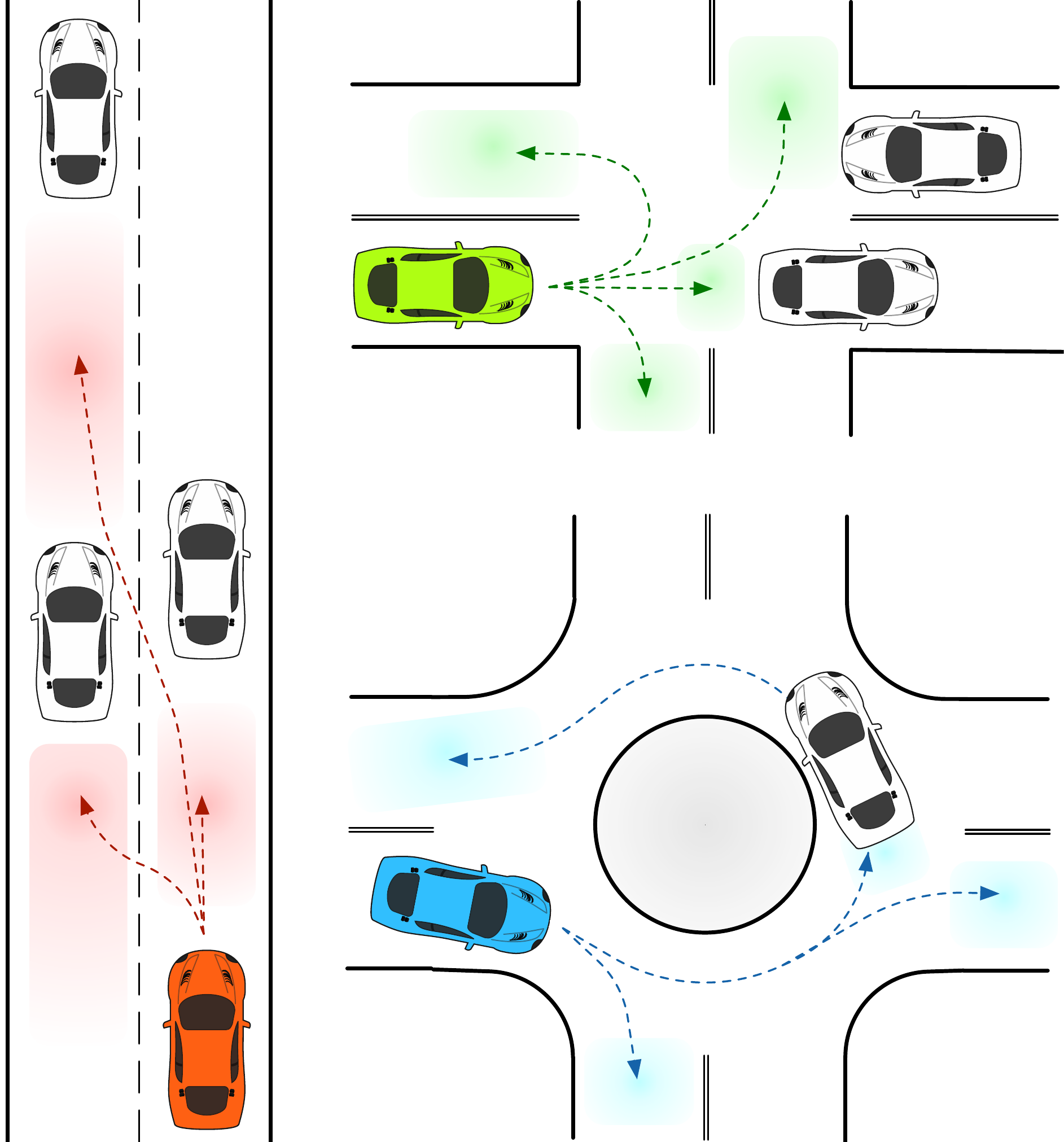}
	\caption{Insertion areas (colored regions) under different driving scenarios for the predicted vehicle.}
	\label{fig:DIA}
\end{figure}

There have been numerous works focused on predicting vehicle behavior which can be divided into two categories: \textbf{intention/maneuver prediction} and \textbf{motion prediction}. Many intention estimation problems have been solved by using classification strategies, such as Support Vector Machine (SVM) \cite{SVM_class}, Bayesian classifier \cite{BF_class}, Hidden Markov Models (HMMs) \cite{HMM_class}, and Multilayer Perceptron (MLP) \cite{MLP_class}. 
Most of these approaches were only designed for one particular scenario associated with limited intentions.
For example, \cite{SVM_class}-\cite{MLP_class} dealt with non-junction segment such as highway, which involves lane keeping (LK), lane change left (LCL) and lane change right (LCR) maneuvers. Whereas \cite{HMM_class}-\cite{IDM} concentrated on junction segment such as intersection, which includes left turn, right turn, and go straight maneuvers. However, in order for autonomous vehicles to drive through dynamically changing traffic scenes in real life, an intention prediction module that can adapt to different scenarios with various possible driving maneuvers is necessary. \cite{insert_2} proposed a maneuver estimation approach for generic traffic scenarios, but the classified driving maneuvers are too specific, which will not only require multiple manually-selected classification thresholds, but also raise problems when unclassified maneuvers occur. 

As a result, we proposed to use semantics to represent the driver intention, which is defined as the intent to enter each \textit{insertion area}. These areas can be the available gaps between any two vehicles on the road or can be the lane entrances/exits. Fig.~\ref{fig:DIA} visualizes the insertion areas under distinct environments. An advantage of using semantic approach is situations can be modeled in a unified way \cite{semantic} such that varying driving scenarios will have no effect on our semantics defined problem. Even for a scenario that has a combination of all the road structures in Fig.~\ref{fig:DIA}, the proposed semantic definition still holds.

Motion prediction is mostly treated as a regression problem, where it tries to forecast the short-term movements and long-term trajectories of vehicles. By incorporating motion prediction with intention estimation, not only the high-level behavioral information, but also the future state of the predicted vehicle can be obtained. For short-term motion prediction, various approaches such as constant acceleration (CA), Intelligent Driver Model (IDM) \cite{IDM}, and Particle Filter (PF) \cite{PF} have been suggested. The main limitation of these works, however, is that they either considered simple cases such as car following or did not take environment information into account. 

For future trajectory estimation, Dynamic Bayesian Networks (DBN) \cite{DBN} and other regression models have been used in several studies. Methods based on artificial neural network (ANN) are also widely applied. In [11], the authors used the LSTM to predict the vehicle trajectory in highway situation. [12] brought forward a Deep Neural Networks (DNN) to obtain the lateral acceleration and longitudinal velocity. However, these approaches only predicted the most likely trajectory for the vehicle without considering uncertainties in the environment. To counter this issue, a Variational Gaussian Mixture Model (VGMM) was proposed for probabilistic long-term motion prediction \cite{VGMM_2012}. Nevertheless, the method was only tested in a simulation environment and the input contains history information over a long period of time, which is usually unaccessible in reality. There are also researches that project the prediction step of a tracking filter forward over time, but the growing uncertainties often cause future positions to end up at some physically impossible locations. 

In contrast, works such as \cite{Ziebart}\cite{E.Rehder} highlighted that by predicting goal locations and assuming that agents navigate toward those locations by following some optimal paths, the accuracy of long-term prediction can be improved. The main advantage of postulating destinations instead of trajectories is that it allows one to represent various dynamics and to automatically incorporate environment constraints for unreachable regions. 

Apart from obtaining the possible goals of predicted vehicles, the required time to reach those locations is also an essential information especially for the subsequent trajectory planning of the ego vehicle. Therefore, many attempts have been made in order to directly predict temporal information. \cite{TTLC_NN_2017} used LSTM to forecase time-to-lane-change (TTLC) of vehicles under highway scenarios. A recent work \cite{TTLC_QRF_2017} utilized the Linear Quantile Regression (LQR) and Quantile Regression Forests (QRF) methods for the probabilistic regression task of TTLC. The authors also concluded that QRF has better performance than LQR.
 
In this paper, Semantic-based Intention and Motion Prediction (SIMP) method is proposed. It utilizes deep neural network to formulate a probabilistic framework which can predict the possible semantic intention and motion of the selected vehicle under various driving scenarios. The introduced semantics for this prediction problem is defined as answering the question of \textit{"Which area will the predicted vehicle most likely insert into? Where and when?"}, which incorporates both the goal position and the time information into each insertion area. Moreover, the adoption of probability can take into account the uncertainty of drivers as well as the evolution of the traffic situations.

The remainder of the paper is organized as follows: Section II provides the concept of the proposed SIMP method; Section III discusses an exemplar scenario to apply SIMP; evaluations and results are provided in Section IV; and Section V concludes the paper.
\section{Concept of Semantic-based Intention and Motion Prediction (SIMP)}

In this section, we first provide a brief overview of Mixture Density Network (MDN), which is an idea we utilize for our proposed method. Then, the detailed formulation and structure of the SIMP method are illustrated.

\subsection{Mixture Density Network (MDN)}

Mixture Density Network is a combination of ANN and mixture density model, which was first introduced by Bishop \cite{MDN_1994}. The mixture density model can be used to estimate the underlying distribution of data, typically by assuming that each data point has some probability under a certain type of distribution. By using a mixture model, more flexibility can be given to model completely general conditional density function $p(\bm{y}|\bm{x})$, where $\bm{x}$ is a set of input features and $\bm{y}$ is a set of output. The probability density of the target data is then represented as a linear combination of kernel functions in the form 
\begin{eqnarray} \label{eq:pdf}
p(\bm{y}|\bm{x}) = \sum_{m=1}^{M}\alpha_m(\bm{x})\phi_m(\bm{y}|\bm{x}),
\end{eqnarray}
where M denotes the total number of mixture components and the parameter $\alpha_m(\bm{x})$ denotes the $m$-th mixing coefficient of the corresponding kernel function $\phi_m(\bm{y}|\bm{x})$.  Although various choices for the kernel function was possible, for this paper, we utilize the Gaussian kernel of the form
\begin{eqnarray} \label{eq:gaussian}
\phi_m(\bm{y}|\bm{x}) = \mathcal{N}(\bm{y} |\mu_{m}(\bm{x}),\sigma_{m}^2(\bm{x})).
\end{eqnarray} 

Such formulation is called the Gaussian Mixture Model (GMM)-based MDN, where a MDN maps input $\bm{x}$ to the parameters of the GMM (mixing coefficient $\alpha_m$, mean $\mu_{m}$, and variance $\sigma_{m}^2$ ), which in turn gives a full probability density function of the output $\bm{y}$. It is important to note that the parameters of the GMM need to satisfy specific conditions in order to be valid: the mixing coefficients $\alpha_{m}$ should be positive and sum to 1; the standard deviation $\sigma_{m}$ should be positive. The use of softmax function and exponential operator in (\ref{eq:constrains}) fulfills the aforementioned constraints. In addition, no extra condition is needed for the mean $\mu_{m}$.
\begin{eqnarray}  \label{eq:constrains}
\alpha_m = \frac{\exp(z_{m}^{\alpha})}{\sum_{i=1}^{M}\exp(z_{i}^{\alpha})},\quad
\sigma_{m} = \exp(z_{m}^{\sigma}), \quad
\mu_{m} = z_{m}^{\mu}
\end{eqnarray}
The parameters $z_{m}^{\alpha}$, $z_{m}^{\sigma}$, $z_{m}^{\mu}$ are the direct outputs of the MDN corresponding to the mixture weight, variance and mean for the $m$-th Gaussian component in the GMM. 

The objective of training the MDN is to minimize the negative log-likelihood as loss function
\begin{eqnarray}
Loss = -\sum_{n}log\bigg\{\sum_{m=1}^{M}\alpha_m^n(\bm{x}^n)\phi_m(\bm{y}^n|\bm{x}^n)\bigg\},
\end{eqnarray}
where $n$ denotes the number of training data. The detailed derivations on closed-form gradient formulation can be found in \cite{MDN_1994}, which demonstrated the capability of training the MDN using back propagation.

\subsection{Proposed SIMP Method}

Our task is to generate probability distributions of the designed semantic description given some representation of the current state. 
We assign a Gaussian Mixture Model (GMM) to each insertion area and multiple GMMs will be involved in one driving scenario. Each Gaussian mixture models the probability distribution of a certain type of motion for the predicted vehicle. Since obtaining the insertion location and the arriving time are the focus of our interests, a 2D Gaussian mixture is used and the predicted variables are constructed as a two dimensional vector: $\bm{y}=[y_{s},y_{t}]^T$. The variable $y_s$ describing the vehicle locations and the variable $y_t$ describing the time information, can be specifically defined according to the driving environment.



Given the current state features $\bm{x}$, the probability distribution $\bm{y}_a$ over a single area $a$ for the predicted vehicle is of the form
\begin{eqnarray}
f(\bm{y}_a|\bm{x}) = \sum_{m=1}^{M}\alpha_{m}\mathcal{N}(\bm{y}_a |\bm\mu_{m},\Sigma_{m})
\end{eqnarray}
with mean and covariance constructed as
\begin{eqnarray}
\bm\mu_{m} = 
\begin{bmatrix} \mu_{s,m} \\ \mu_{t,m} \end{bmatrix}, \quad
\Sigma_m = \begin{bmatrix} \sigma_{s,m}^2 & \rho_{m}\sigma_{s,m}\sigma_{t,m} \\ \rho_{m}\sigma_{s,m}\sigma_{t,m} & \sigma_{t,m}^2 \end{bmatrix},
\end{eqnarray}
where $\rho_{m}\in \left[-1,1\right]$ is the correlation coefficient.

In addition to formulate a regression model for each insertion area, we also require the probability of entering each area for the predicted vehicle. Therefore, Deep Neural Network (DNN) was used as the basis for our Semantic-based Intention and Motion Prediction (SIMP) structure. The output of the network contains both necessary parameters for every 2D Gaussian Mixture Model (GMM) and the weight $w_a$ for each insertion area $a$. 

For the desired outputs, we expect not only the largest weight to be associated to the actual inserted area, but also the highest probability at the correct location and time for the output distributions of that area. Consequently, we define our loss function as
\begin{equation} \label{eq:loss}
\begin{split}
L &= W_1\bigg(-\sum_{n}log\bigg\{\sum_{a=1}^{N_a}\hat{w}_{a}^n f(\bm{y}_a^n|\bm{x})\bigg\}\bigg) \\
  & \quad +W_2\bigg(-\sum_{n}\sum_{a=1}^{N_a}\hat{w}_{a}^nlog(w_{a}^n)\bigg),
\end{split} 
\end{equation}

\begin{figure}[htbp]
	\centering
	\includegraphics[scale=0.5]{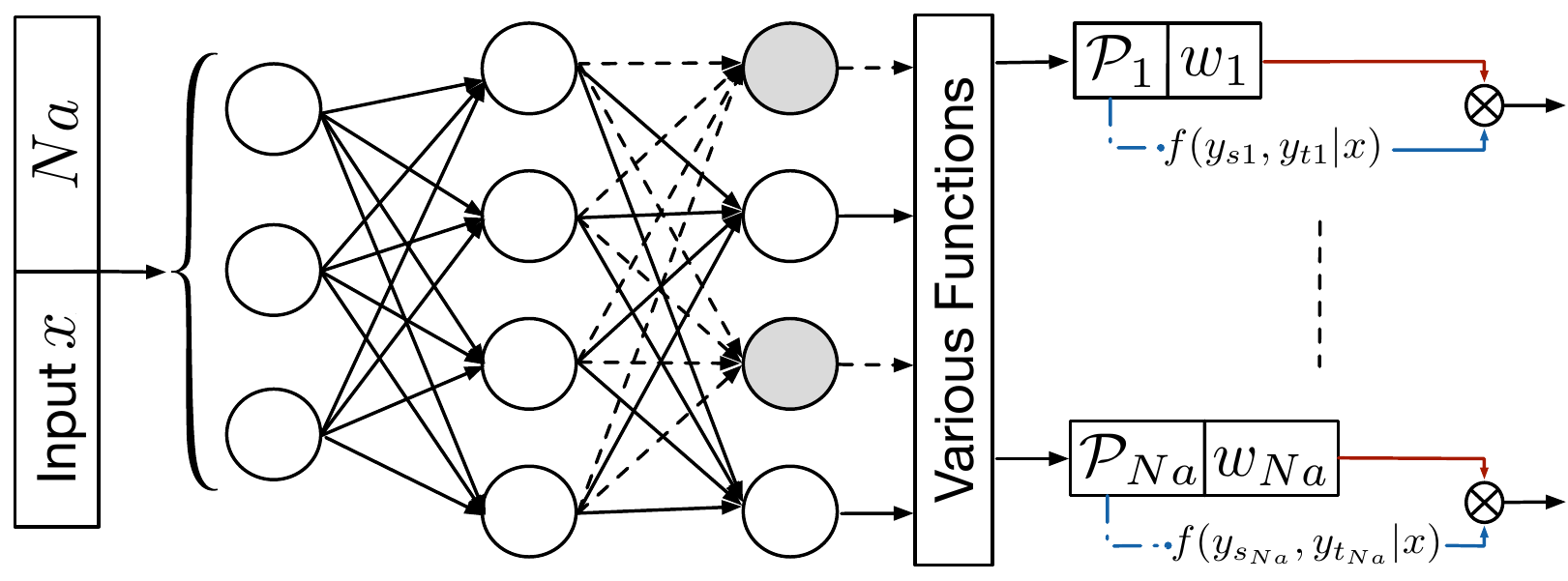}
	\caption{Structure of the SIMP Method }
	\label{fig:SIMP_structure}
\end{figure}
where $N_a$ denotes the total number of insertion areas and $\hat{w}_{a}$ denotes the ground truth, which is the one-hot-encoding of the final area that the predicted vehicle entered. The last term denotes the cross-entropy loss of the area weights. Parameters $W1$ and $W2$ need to be manually tuned such that the two loss components will have the same order of magnitude during training.

The overall architecture of our SIMP method is shown in Fig.~\ref{fig:SIMP_structure}. Due to the first-order Markov assumption, the input features  depend only on the current time step. The network consists of an input layer, several fully connected layers, and a dropout layer which ensures better generalization and prevents overfitting of the training data. After passing different types of parameters through corresponding functions, the output will satisfy the aforementioned constraints. For $N_a$ insertion areas, the total number of output parameters can be calculated as: $N_a*(M*6 + 1)$. The interpretation is: there is a weight parameter $w_a$ associated to each area $a\in N_a$, and for every $m\in M$ within an area, six parameters are needed, $\mathcal{P}_a^m =$ \{$\alpha_{m},\mu_{s,m},\mu_{t,m},\sigma_{s,m},\sigma_{t,m},\rho_{m}$\}, to formulate the 2D GMM.

\section{An Exemplar Highway Scenario}
In this section, we use an exemplar highway scenario to apply the proposed Semantic-based Intention and Motion Prediction (SIMP) method. The data source and detailed problem formulation are presented. 
\subsection{Dataset}
All the data we used was taken from the NGSIM US 101 dataset which is publicly available online at \cite{NGSIM}. It contains detailed vehicle trajectory data collected on the highway with 10 Hz sampling frequency. The measurement area is approximately 640 meters (2100 feet) in length and there are five freeway lanes plus an auxiliary lane for the on/off-ramp.
For each vehicle that performs a lane change maneuver, we picked up to 40 frames (4s) before the vehicle's center intersects the lane mark; for vehicles that keep driving on the same lane for a long period, we considered these frames as input for the lane keeping maneuver. A total of 17,179 frames were selected from the dataset and splitted into 80\% for training and 20\% for testing.

\subsection{Scenario and Problem Description}
\begin{figure}[htbp]
	\includegraphics[scale=0.45]{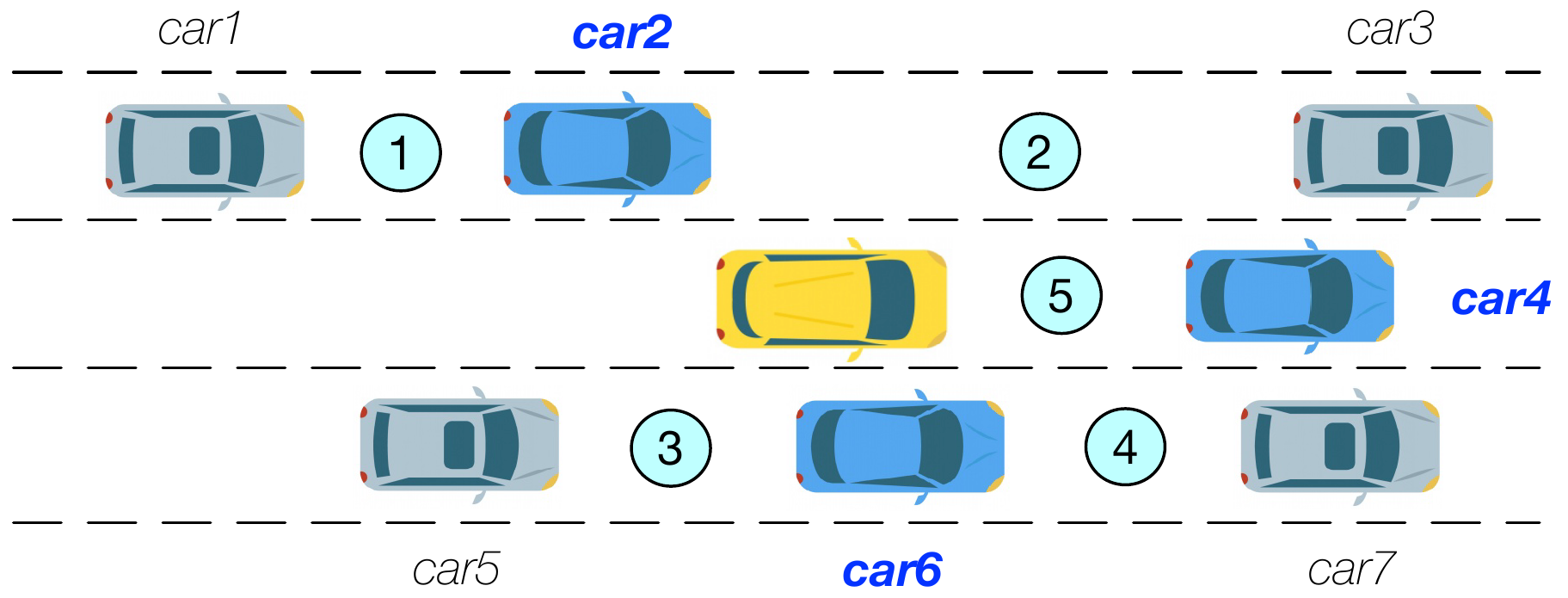}
	\caption{An exemplar driving scenario}
	\label{fig:fake_scenario}
\end{figure}

A representation of the exemplar highway driving scenario is shown in Fig.~\ref{fig:fake_scenario}. The yellow car is the vehicle we decide to predict; the three blue cars (\textit{car2, car4}, and \textit{car6}) are the reference vehicles, which are selected as
having the closest Euclidean distance to the predicted vehicle on each of the three lanes (we consider only the front vehicle on predicted vehicle's lane); the four gray cars (\textit{car1, car3, car5}, and \textit{car7}) are named as `other vehicles', which are vehicles in front and behind each of the two reference cars: \textit{car2} and \textit{car6}. If any of these surrounding vehicles is too far from the predicted vehicle, we consider it as nonexistence within the range of the current scenario. Therefore, for each input frame, a maximum of three driving lanes and seven vehicles are considered.

In Fig.~\ref{fig:fake_scenario}, there are five circled areas that our predicted vehicle could end up going into and we name them as Dynamic Insertion Area (DIA). If the predicted vehicle (yellow car) inserts into area 1-4, a lane change behavior is indicated; however, if it inserts into area 5, a lane keeping behavior is implied. These areas are dynamic because both their locations and sizes will vary at each time step. 

In this particular highway scenario setting, the output $y_s$ represents the absolute distance between the final insertion point and the corresponding reference vehicle for that inserted area; $y_t$ represents the time-to-lane-change (TTLC) of the predicted vehicle. When the center of the vehicle intersects the lane mark, TTLC = 0. For the lane keeping situation, TTLC is set to a large number (4s) to represent that the vehicle has not yet decided to change the lane.

\subsection{Features and Structure Details}  
For each input frame, a total of 25 input features are selected which are listed in Table~\ref{tab:feature}. Each input frame corresponds to 3 types of labels extracted from data: area weight, final goal location, and remaining insertion time. According to the data, the longitudinal direction is the driving direction. The current lane center (CLC) denotes the midpoint of the current lane. Because of the small angle difference between the front and the predicted vehicle, only the relative angle information for the left and right reference vehicles are considered.
Time-to-collision (TTC) is calculated by dividing the speed difference by the relative distance of two vehicles. We compute the inverse of time-to-collision (iTTC) instead due to the existence of infinity TTC value as the speed difference gets close to zero.

As mentioned previously, there will be a maximum of seven cars within each input frame. If, however, a vehicle does not exist, we assign its longitudinal distance to a large number and its velocity to be the same as that of the predicted vehicle. If there is no available lane on one side of the predicted vehicle, we set the three vehicles in that nonexistent lane to be close to each other and the reference vehicle to be directly above/below the predicted vehicle. Similarly, all these three vehicles are set to have the same speed as the predicted vehicle. Such setting can guarantee the feasibility of the predicted results.

As for the network structure, we use three fully connected layers of 400 neurons each, with $tanh$ non-linear activation function. After that, a dropout layer of rate 0.5 is appended. The parameter $N_a$ is five for this particular scenario.
\begin{table}[ht]
	\caption{Features for One Input Frame}
	\label{tab:feature}
	\centering
		\begin{tabular}{c m{1.25cm} p{5cm}}
			\toprule 
			& \textit{Feature} & \textit{Description} \\ 
			\midrule \midrule
			\multirow{2}{*}[-0.1cm]{\shortstack[lb]{\textbf{Predicted} \\ \textbf{Vehicle}}}  & $v_{pred}^{y}$ & Absolute velocity in longitudinal direction\\ [1.2mm]
			& $d_{CLC_{pred}}^{x}$ & Lateral distance to the current lane center \\ 
			\midrule
			\multirow{5}{*}[-0.8cm]{\shortstack[lb]{\textbf{Reference} \\ \textbf{Vehicles}}} & $v_{ref}^{y}$ & Absolute velocity in longitudinal direction \\ [1.2mm]
			& $d_{ref,pred}^{y}$ & Position in longitudinal direction, relative to predicted vehicle \\
			& $d_{(l,r),pred}^{x}$ & Relative lateral position between left/right reference vehicle and predicted vehicle \\
			& $\theta_{(l,r),pred}$ & Relative angle between left/right reference vehicle and predicted vehicle \\ 
			& $iTTC_{f,pred}$ & Inverse time-to-collision between front reference vehicle and predicted vehicle \\
			\midrule
			\multirow{3}{*}[-0.3cm]{\shortstack[lb]{\textbf{Other} \\ \textbf{Vehicles}}}  & $v_{o}^{y}$ & Absolute velocity in longitudinal direction \\ [0.7mm]
			& $d_{o,pred}^{y}$ & Position in longitudinal direction, relative to predicted vehicle \\
			& $iTTC_{o,ref}$ & Inverse time-to-collision relative to corresponding reference vehicle \\ 
			\bottomrule
		\end{tabular}
\end{table}
\section{Evaluation and Results}
In this section, different evaluation techniques are presented to assess the model quality and the final results are discussed. 
\subsection{Evaluation Setup}
\subsubsection{Baseline Model}
To evaluate our SIMP method, we trained a Support Vector Machine (SVM) \cite{SVM} and a Quantile Regression Forests (QRF) \cite{QRF} separately. Since SVM is wildly used for classification problems, we compared it with the intention prediction part of our framework. The QRF is a combination of Quantile Regression and Random Forests \cite{RF}, which extends the concept of tree ensemble learning to probabilistic prediction. Instead of point estimating the conditional mean for the selected variables like other regression methods, the objective is to estimate an arbitrary conditional quantile. The quantiles can provide detailed information of the minimum and maximum values for the dependent variable and encompass the uncertainty estimation. Hence, we compared our motion prediction part of the probabilistic framework with the QRF method for evaluation. The details of the baseline models are presented below
\begin{figure*}[htbp]
	\centering
	\includegraphics[scale=0.19]{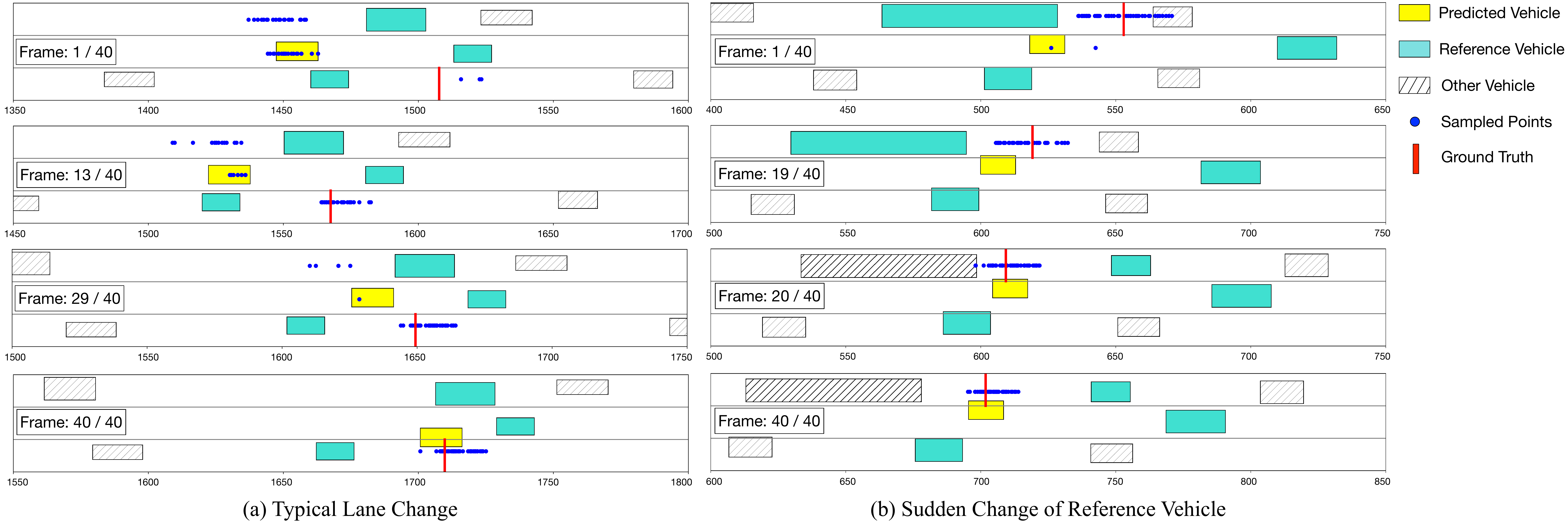}
	\caption{Two example cases to visualize the performance. In each testing frame, 50 points were sampled by two steps: 1{)} multiply the total number of dots by each DIA weight. 2{)} for every dot assigned to each DIA, sample it according to the corresponding distribution of that area. \textit{(The unit of the horizontal axis is in feet.)}}
	\label{fig:example_scene}
\end{figure*}
\begin{itemize}
	\item SVM: kernel =  (Gaussian) radial basis function (RBF)
	\item QRF: ntree = 1000, mtry = 5, nodesize = 10
\end{itemize}
where \textit{ntree} is the number of trees in the forest, \textit{mtry} is the number of random features in each tree, and \textit{nodesize} is the minimal size of terminal nodes. All these parameters were selected using five-fold cross validation.

\subsubsection{Evaluation for Intention Estimation}
For training and testing, each sample from our data was assigned to a semantic intention class, which is expressed as $I\in \{area1, area2, area3, area4, area5\}$. However, since these dynamic insertion areas (DIA) change constantly during the driving period, it is hard to detect the final insertion area at the early stage. Therefore, for better evaluation at the beginning of the input driving segments, we merged the original five semantic intentions into three: $\{LCL,LCR,LK\}$, where $\{area1,area2\} \in LCL$, $\{area3,area4\}\in LCR$, and $\{area5\}\in LK$. During training, the input features for SVM were the same as our method, and the labels were the corresponding final DIA numbers. The evaluation contains three steps: 
\renewcommand{\labelenumi}{\roman{enumi}.}
\begin{enumerate}
	\item For all testing data, create the Receiver Operating Characteristic (ROC) curve to compare our method with SVM. (Use the simplified 3 intention classes.)
	\item Find the best threshold from the ROC curve and use it to calculate the \textit{recall, precision, F1 score} as well as the \textit{average prediction time} for both methods.
	\item For testing data that has a TTLC smaller than the obtained \textit{average prediction time}, analyze the performance of each DIA. (Use the original 5 semantic intention classes.)
\end{enumerate}

\subsubsection{Evaluation for Motion Prediction}
In our problem setting, two semantic described motions are predicted: final locations in each insertion area (destination) and the remaining time to reach those locations (TTLC). 
For the conditional distribution of each motion, we expect not only small d{\tiny {\tiny }}ifference between the predicted mean and the actual value, but also centralized distribution around the predicted mean. Hence, we evaluated the \textit{root mean squared error} (RMSE) of the output mean as well as the confidence interval for both the QRF and the SIMP method. The number of mixture components $M$ for each DIA was set to one for analysis purpose. For the training process of QRF, we trained two separate random forest quantile regressors, where the input features remains the same and the label is either the location or the time information.

Two different intervals were chosen to assess the testing results for each method:
\begin{itemize}
	\item SIMP-1$\sigma$: one standard deviation interval 
	\item SIMP-2$\sigma$: two standard deviation interval
	\item QRF-68\%: 16\% to 84\% quantile interval
	\item QRF-95\%: 2.5\% to 97.5\% quantile interval.
\end{itemize}

\subsection{Results and Discussion}

\subsubsection{Visualization of Selected Cases}
We selected two distinct traffic situations to visualize our results. Each situation had 40 frames (4s) and we chose four representative frames from each case to illustrate the overall performance. The full video can be found on \url{https://www.youtube.com/watch?v=6A3Hl-mRhbI}. 

\begin{figure}[htbp]
	\centering
	\includegraphics[scale=0.4]{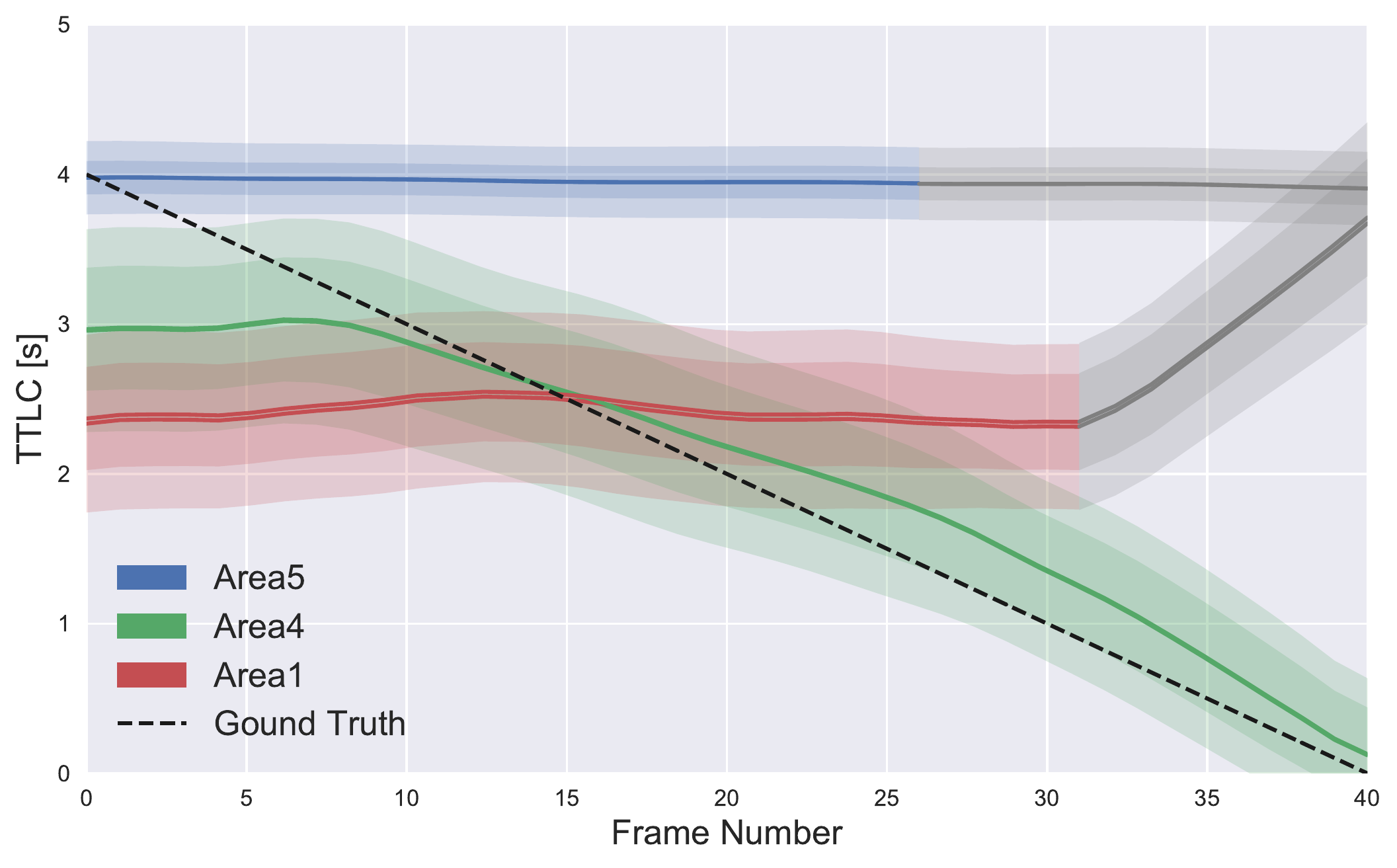}
	\caption{TTLC illustration for the case in Fig.~\ref{fig:example_scene}(a). We sampled 100 points from the mixture distribution of each related DIA and plotted the mean as well as the 3$\sigma$ and 1$\sigma$ prediction intervals for these samples. When area weight is too small to be associated with sampled points, the TTLC result of that area at the corresponding frame will be colored in gray.}
	\label{fig:TTLC_LC}
\end{figure}

A typical lane change situation is illustrated in Fig.~\ref{fig:example_scene}(a) where the sampled points are all in the proper DIA for each frame. It is reasonable to have several possible areas at the early stage since there are multiple choices for the driver and no specific one has been chosen yet. It should be note that it is difficult to numerically justify the correctness of these circumstances without using the human-labeled ground truth. However, as soon as the driver decides where to go, our result could be compared with the label extracted from data. We further used this case to illustrate the TTLC prediction result in Fig.~\ref{fig:TTLC_LC}. The differences between our resulted mean and the ground truth are all smaller than 0.3s within three seconds before lane change; besides, the predicted TTLC values for other insertion areas remain in reasonable ranges.


Since the reference vehicle will switch from one to another while the predicted vehicle is driving, we need to guarantee the capability of our method to handle such cases without large discontinuity on the prediction result. Therefore, we examined one of such cases shown in Fig.~\ref{fig:example_scene}(b) and it can be observed that such sudden change occurs between frame 19 and 20. During this period, our sampled points are able to keep in the correct DIA and tightly distributed around the red target line.




\subsubsection{Intention Estimation}
\begin{figure}[htbp]
	\centering
	\includegraphics[scale=0.45]{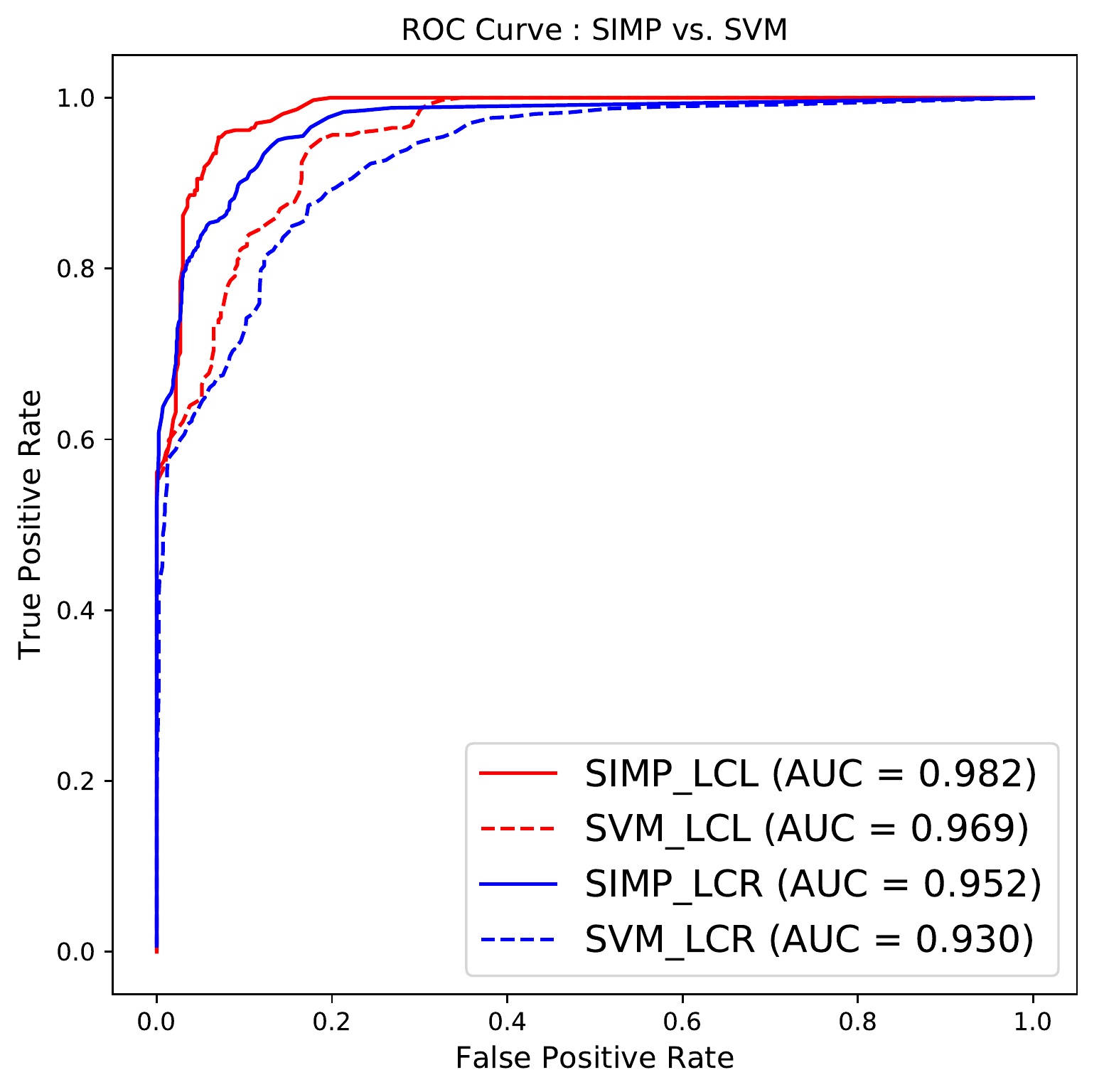}
	\caption{ROC curve comparison}
	\label{fig:ROC_compare_SVM}
\end{figure}

The ROC curves of the SIMP and the SVM methods are visualized in Fig.~\ref{fig:ROC_compare_SVM}. The curves were created by plotting the \textit{true positive rate} (TPR) against the \textit{false positive rate} (FPR) at various threshold settings. Similar to \cite{TTLC_NN_2017}, we defined two positive classes (lane change left and right) and one negative class (lane keeping). The area under the ROC curve (AUC) can be used as an aggregated measure of the classifier performance. The true positive (TP) represents correct prediction of either lane change left or right, the false positive (FP) indicates mispredicting the lane change direction, and the false negative (FN) means incorrectly predicting a lane change into lane keeping. 

From Fig.~\ref{fig:ROC_compare_SVM} and AUC values, we observe that our method outperforms SVM for lane change maneuvers. A threshold of 0.3 for classification was chosen for making the best trade-off between a high TPR and a low FPR. Given the selected threshold, we can further calculate the \textit{precision} and \textit{recall} as
\begin{eqnarray}
precision = \dfrac{TP}{TP + FP}, \quad recall = \dfrac{TP}{TP + FN}
\end{eqnarray}
and the F1 score can be obtained by the formula
\begin{eqnarray}
F1 = \dfrac{2*precision*recall}{precision + recall},
\end{eqnarray}
which denotes how good the classification abilities are. Moreover, how early the lane change can be recognized is also in the focus of our interests. Thus, we calculated the average prediction time from the testing data that were classified as true positive. The overall performance of the two methods are compared in Table~\ref{tab:comparison}. It is apparent from table that the proposed method has better performance than SVM in terms of both prediction accuracy and the average prediction time.

Since our method can correctly forecast the predicted vehicle's intention approximately 2s in advance to the actual lane change according to Table~\ref{tab:comparison},  we further plotted the ROC curve and calculated the AUC for each dynamic insertions area (DIA) to examine how well can SIMP predict the final insertion region. The obtained AUC values for \textit{Area1, Area2,} and \textit{Area3} are all equal to 1, and \textit{Area4} has a 0.994 AUC value. The result implies that the proposed method can not only detect the lane change direction but also the specific dynamic insertion area (DIA) with high accuracy for the selected time window.

\begin{table}[ht]
	\caption{Performance Comparison}
	\label{tab:comparison}
	\centering
	\begin{tabular}{p{1.25cm} p{1cm} p{1cm} p{1cm} p{1.5cm}}
		\toprule 
		Method & Precision & Recall & F1-Score & Avg. Predict Time (s) \\
		\midrule \midrule
		SVM & 0.859 & 0.919 & 0.888 & 1.911 \\
		\midrule
		SIMPF & 0.936 & 0.925 & 0.931 & 1.957 \\
		\bottomrule
	\end{tabular}
\end{table}

\subsubsection{Motion Prediction}
\begin{figure}[htbp]
	\centering
	\includegraphics[scale=0.39]{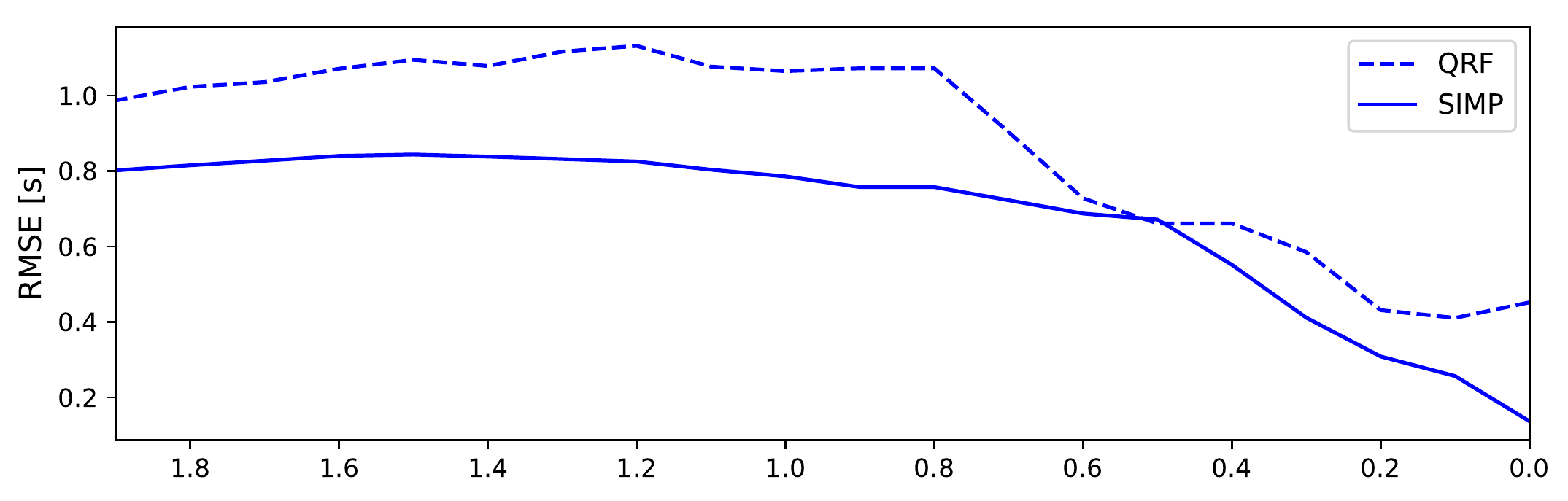}
	\includegraphics[scale=0.39]{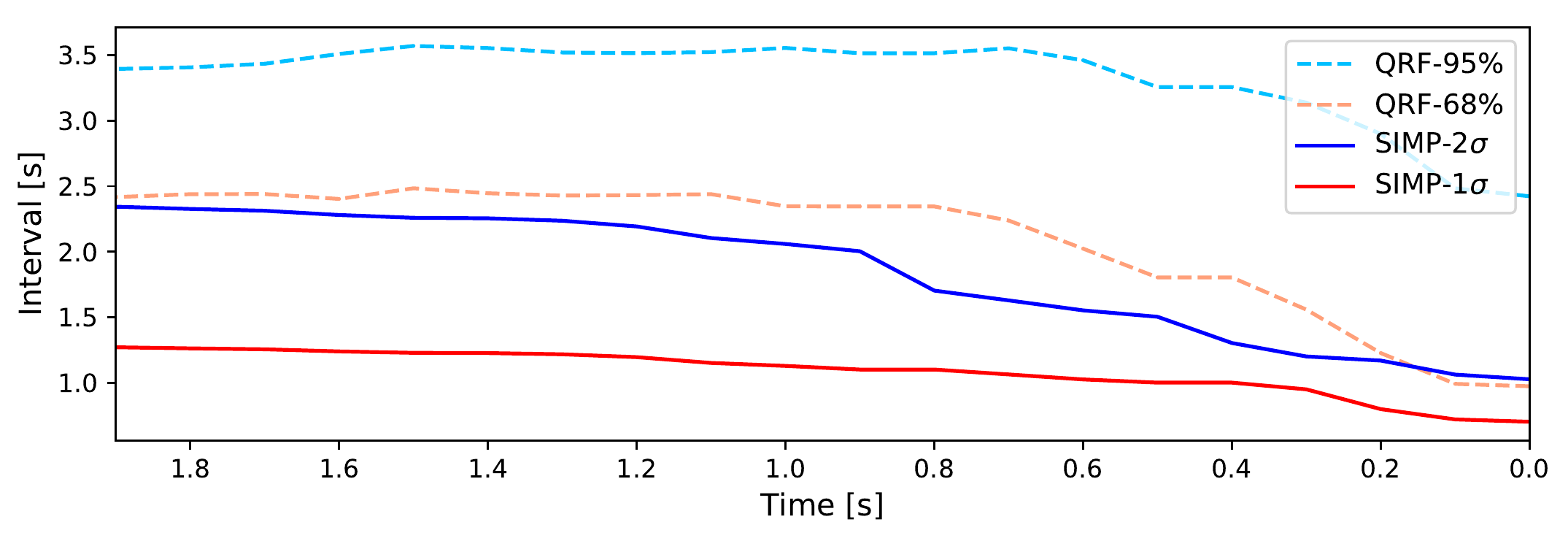}
	\caption{Comparison of Time-to-Lane-Change (TTLC) Prediction}
	\label{fig:ttlc_mean_compare}
\end{figure}

\begin{figure}[htbp]
	\centering
	\includegraphics[scale=0.39]{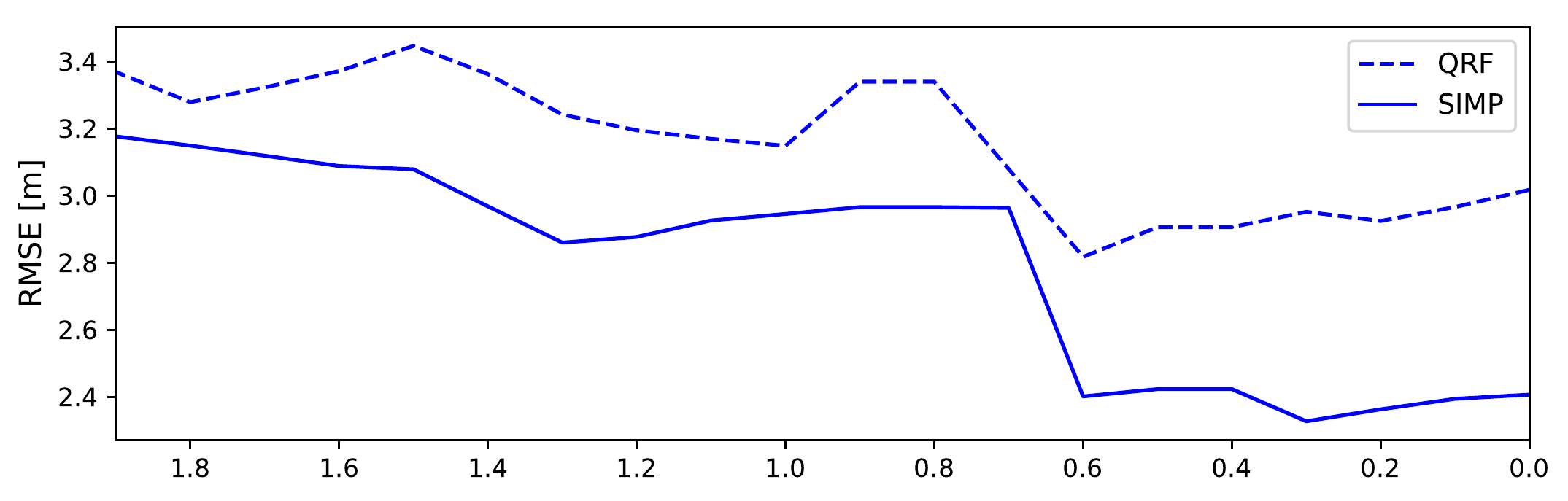}
	\includegraphics[scale=0.39]{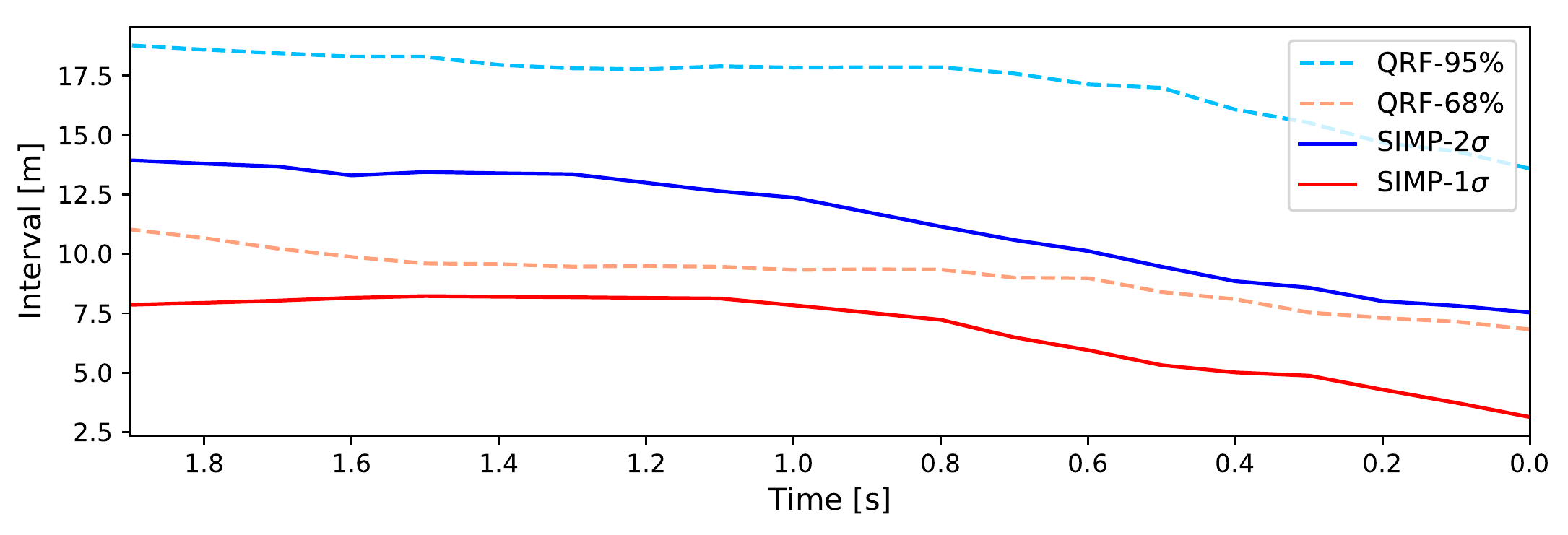}
	\caption{Comparison of Destination Prediction}
	\label{fig:dist_mean_compare}
\end{figure}

The comparison results between QRF and the proposed method for two motion prediction tasks are shown in Fig.~\ref{fig:ttlc_mean_compare} and Fig.~\ref{fig:dist_mean_compare}. The mean for QRF was obtained by calculating the 50\% quantile (or median) assuming symmetric distribution. We utilized the testing data that has a TTLC smaller than the average prediction time derived in the previous section. The mean and confidence interval were calculated from the obtained output distribution of the correct insertion area. As can be seen in the plots, the RMSE of our approach for both motion predictions are smaller compared with the QRF method. The RMSE error of the TTLC tends towards zero for the lane change cases by using the SIMP method. One thing need to mention is that the error for the destination prediction is not close to zero even at $t=0$. However, this is not unexpected given the fact that the predicted distance is relative to the reference car. Thus, the results might deviate due to the consideration of any velocity variance of the reference vehicle.

For the confidence interval comparison, it is obvious to see that the performance of our proposed method surpasses QRF especially for the TTLC prediction, where the 2-$\sigma$ interval of the SIMP method is even smaller than the 68\% interval of the QRF. The gradually decreasing difference between the one and two standard deviation interval as well as the declining interval values imply that our predicted Gaussian distribution is becoming more centralized around the ground truth as the TTLC approaching zero.

\section{Conclusions}
In this paper, a Semantic-based Intention and Motion Prediction (SIMP) method was proposed, which can generate various designated conditional distributions for predicted vehicles under any circumstances. An exemplar highway scenario with real-world data was used to apply the idea of SIMP. First, two representative driving cases were utilized to visualize the testing result. Then the intention prediction and the motion prediction part were separately compared with two different baseline models: SVM and QRF. Our approach outperforms these methods in terms of both the prediction error and the confidence intervals. The key conclusion is that by combining different prediction tasks using semantics in a single framework, we can not only easily generalize the idea into any traffic scenarios but also obtain competitive performance compared to traditional methods. The output goal position and time information can be further used to generate optimal trajectories for predicted vehicles and eventually obtain a desirable path for our own autonomous vehicle. For future work, we will examine the SIMP method on more complex scenarios as well as take into account the occurrence of vehicle occlusion.

\addtolength{\textheight}{-12cm}   




\end{document}